\documentclass[conference]{IEEEtran}
\IEEEoverridecommandlockouts

\usepackage{cite}
\usepackage{amsmath,amssymb}
\usepackage{graphicx}
\usepackage{booktabs}
\usepackage{array}
\usepackage{xcolor}
\usepackage{url}
\usepackage{soul}

\newif\ifshowrevisions
\showrevisionsfalse
\sethlcolor{yellow}
\ifshowrevisions
  \newcommand{\rev}[1]{\hl{#1}}
\else
  \newcommand{\rev}[1]{#1}
\fi

\newcommand{\Full}{\textsc{Full}}
\newcommand{\Skip}{\textsc{Skip}}
\newcommand{\Neut}{\textsc{Neutralize}}
\newcommand{\act}[1]{\textit{#1}}

\title{D'BEAST: Logged Bad Experiences as Explanation for Autonomous Driving Agents}

\author{
\IEEEauthorblockN{
Ahmad Alfan Alfian Irfan\textsuperscript{1},
Nur Ahmad Khatim\textsuperscript{2},
Mansur Arief\textsuperscript{3}
}
\thanks{\textsuperscript{1}Information Technology,
Universitas Muhammadiyah Yogyakarta,
Yogyakarta, Indonesia.
Email: ahmad.alfan.ft23@mail.umy.ac.id}
\thanks{\textsuperscript{2}Division of Information Science,
Nara Institute of Science and Technology,
Japan.
Email: ahmad.khatim\_nur.ak9@naist.ac.jp}
\thanks{\textsuperscript{3}Department of Industrial and
Systems Engineering, IRC Smart Mobility and Logistics,
King Fahd University of Petroleum and Minerals
(KFUPM), Saudi Arabia.
Email: mansur.arief@kfupm.edu.sa}
}

\begin{document}
\maketitle

\begin{abstract}
Reinforcement learning agents learn from mistakes, yet their decisions can be difficult to \rev{trace back} to those experiences.
We \rev{investigate} whether a tabular agent's decisions can be explained by its bad experiences, \rev{defined as} the training transitions associated with safety or task events.
\rev{A} fixed training log \rev{is replayed} from the same initial Q-table under three conditions.
\Full{} retains all updates, \Skip{} omits bad-experience updates, and \Neut{} retains those updates without their event penalties.
We illustrate the method \rev{on} six classic Markov decision processes and \rev{evaluate} it in closed-loop \rev{driving in CARLA}.
On the full driving task, \Full{} reached the goal from 33 of 40 test starts, whereas \Skip{} and \Neut{} reached it from none, \rev{indicating that} the bad experiences carry decision-relevant information.
\rev{At the local level, for} 47 decisions shortly before a failure of \Skip{} or \Neut{}, removing or neutralizing a small group of bad \rev{experiences, with a median size of one, caused} the trained agent \rev{to select} the failing policy's action in 46 cases.
For this agent, the results indicate that its decisions can be explained by its bad experiences.
\end{abstract}

\begin{IEEEkeywords}
explainable reinforcement learning, contrastive explanation, training-data attribution, autonomous driving, CARLA
\end{IEEEkeywords}

\section{Introduction}\label{sec:intro}

Reinforcement learning (RL) agents learn sequential decisions through interaction~\cite{mnih2015human} and \rev{have been} studied for lane keeping, intersection handling and motion planning in intelligent vehicles~\cite{kiran2022survey,ye2021survey}.
Explaining \rev{such} learned decisions is \rev{an important} concern in automated driving~\cite{kuznietsov2024explainable}\rev{, and explainable} RL (XRL) examines what an agent has learned~\cite{milani2024survey,vouros2023survey}.

\rev{Existing} XRL methods explain decisions through state features~\cite{greydanus2018visualizing}, reward components~\cite{juozapaitis2019decomposition}, policy summaries~\cite{amir2018highlights}, or causal and counterfactual models~\cite{madumal2020causal,gajcin2024counterfactual}.
Training-data attribution\rev{, by contrast, identifies} which trajectories~\cite{deshmukh2023trajectories}, transitions~\cite{gottesman2020influential}, buffer records~\cite{hu2025snapshot} or individual experiences~\cite{hiraoka2025influential} influence a decision.

Consider a trained agent that, \rev{when} approaching a junction, brakes and steers left rather than driving straight. Did it learn this \rev{behavior} from an earlier straight run that left the road, from the penalty that followed, or from both?
In this work\rev{,} we ask whether an agent's decisions can be explained by its bad experiences.
We pose the contrastive question ``why did the agent take action~A (the fact) rather than action~B (the foil)?''~\cite{miller2019explanation} and identify \rev{the} recorded events on which that preference depends.
A global explanation first checks whether driving changes without these experiences, \rev{a} prerequisite for using them as explanations. \rev{A} local explanation \rev{identifies} the recorded events behind a choice, which a developer can inspect.

The example also \rev{raises the question of} whether the preference rests on the recorded transition, on its event penalty, or on both. Removing an experience as a whole does not distinguish \rev{between} these contributions.
We \rev{therefore} compare the trained agent (\Full) with a replay that omits bad-experience updates (\Skip) and a replay that removes only their event penalties (\Neut), conditional on the same initial table and training log.
Because the replay of a fixed log is exact, both explanations are computed rather than estimated, and for tabular agents they \rev{provide} an exact reference for methods that estimate the effect of removing training data.
To the best of our knowledge, the effect of a hazardous experience has not previously been separated from the effect of its penalty to explain the decisions of an RL driving agent.

Our contributions are threefold. First, we propose logge\textbf{D} \textbf{B}ad \textbf{E}xperiences \textbf{AS} explana\textbf{T}ion (\textbf{D'BEAST}), which takes bad experiences as the unit of explanation for an RL agent's decisions and \rev{distinguishes} omitting a bad experience from removing only its event penalty by exact replay of the training log.
Second, we define a global and a local explanation and test them for foil reproduction, specificity and behavioral relevance, including a closed-loop handover at the explained decision.
Third, we illustrate the method on six classic Markov decision processes (MDPs) and \rev{evaluate} it on a tabular driving agent trained through a curriculum in CARLA~\cite{dosovitskiy2017carla}.

This paper is organized as follows. Sec.~\ref{sec:related} \rev{reviews related work on} explainable RL and training-data attribution.
Sec.~\ref{sec:prelim} \rev{describes} the agent and its training log, and Sec.~\ref{sec:method} presents \rev{the} replay method and the explanation tests.
Sec.~\ref{sec:experiments} \rev{reports the} experiments and \rev{results,} Sec.~\ref{sec:discussion} \rev{discusses} the implications and limitations of \rev{this work, and} Sec.~\ref{sec:conclusion} \rev{concludes the paper}.

\begin{figure*}[t]
\centering
\includegraphics[width=\textwidth]{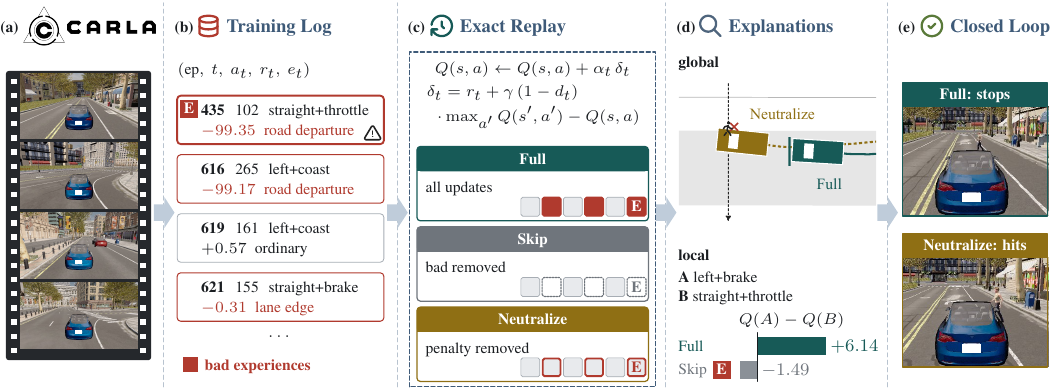}
\caption{Overview of our framework. (a)~Driving in CARLA. (b)~Training log with bad experiences in red, including the running example E. (c)~Exact replay yields \Full, \Skip{} and \Neut. (d)~Global explanation at a pedestrian crossing and local explanation of E by $Q(\mathrm{A})-Q(\mathrm{B})$ under \Full{} and with E skipped. (e)~CARLA frames from one gate start.}
\label{fig:framework}
\end{figure*}

\section{Related Work}\label{sec:related}

\rev{This section reviews explanation methods for RL agents, training-data attribution, and penalties and reward counterfactuals in safe RL.}

\subsection{Explanations for RL Agents}

Among experience-based explanations, Cruz et al.~\cite{cruz2023explainable} explain goal-directed actions using success probabilities estimated from an agent's experience.
In driving, latent representations~\cite{chen2022interpretable} and decision-tree policies~\cite{schmidt2021trust} \rev{have been used to} make learned policies interpretable.
\rev{In contrast, we} use the contrastive form of explanation~\cite{miller2019explanation} to connect an action preference to \rev{specific} recorded safety or task events.

\subsection{Training-Data Attribution}

Training-data attribution measures how training samples contribute to predictions or model performance~\cite{hammoudeh2024training}.
Influence functions~\cite{koh2017influence}, gradient tracing~\cite{pruthi2020tracin} and Data Shapley~\cite{ghorbani2019shapley} estimate these contributions without exhaustive retraining.
In RL, Gottesman et al.~\cite{gottesman2020influential} identify transitions whose removal changes an off-policy value \rev{estimate, and} Deshmukh et al.~\cite{deshmukh2023trajectories} attribute decisions to trajectory clusters, including hazardous trajectories, by retraining an offline agent without each cluster.
Hu et al.~\cite{hu2025snapshot} attribute PPO actions to buffer records through gradient \rev{similarity, and} Hiraoka et al.~\cite{hiraoka2025influential} estimate the influence of \rev{individual} experiences and remove harmful ones to improve the agent.
\rev{Apart from} retraining~\cite{deshmukh2023trajectories}, these methods estimate the effect of \rev{retaining} or removing a sample as a \rev{whole. In contrast,} we replay a fixed log exactly and can \rev{therefore} remove \rev{either} an update or only its event penalty.

\subsection{Penalties and Reward Counterfactuals}

Safe RL often encodes hazards as penalties or constraints~\cite{garcia2015safe, arief2026dijkstra}, and human oversight can block catastrophic actions during training~\cite{saunders2018trial}. We ask instead \rev{to which} of these penalized experiences a trained policy's decisions can be \rev{traced}.
Reward-based counterfactuals change the reward rather than the training data.
Liu et al.~\cite{liu2023counterfactual} compare policies learned under different reward configurations, and Alabdulkarim et al.~\cite{alabdulkarim2025experiential} train influence predictors for each reward source alongside the agent.
Reward decomposition~\cite{juozapaitis2019decomposition} \rev{quantifies} how much all penalties together contribute to an action preference. Our \Neut{} condition attributes \rev{this} contribution to \rev{specific} logged events by changing only their rewards.

\section{Preliminaries}\label{sec:prelim}

We consider a finite MDP with states $s$, actions $a$, rewards $r$, discount $\gamma$ and a terminal flag $d$.
A tabular agent learns action values by Q-learning~\cite{watkins1992qlearning},
\begin{align}
Q(s_t,a_t) \leftarrow{}& Q(s_t,a_t) + \alpha_t\bigl[r_t + \gamma(1-d_t)\max_{a'}Q(s_{t+1},a') \nonumber\\
&- Q(s_t,a_t)\bigr], \label{eq:qlearning}
\end{align}
with a polynomial step size~\cite{evendar2003learning}, $\alpha_t = n_t(s_t,a_t)^{-0.6}$, where $n_t(s_t,a_t)$ counts the updates of the \rev{state--action} pair up to and including update $t$. A terminal transition ($d_t=1$) does not bootstrap, whereas the last transition of a truncated episode \rev{retains} $d_t=0$.
Each logged update $t=1,\dots,T$ stores the transition $(s_t,a_t,r_t,s_{t+1},d_t)$, the applied step size $\alpha_t$, the training episode and step, the source of the action (the learner or, in driving, a teacher) and an event label $e_t$.
Evaluation is greedy with respect to $Q$, with ties broken by \rev{selecting} the first action in a fixed ordering.

\section{Bad Experiences as Explanations}\label{sec:method}

\rev{The proposed} method records \rev{all} training updates and replays them to compare decisions with and without bad experiences (Fig.~\ref{fig:framework}). The global explanation asks whether the bad experiences matter, and the local explanation asks which of them matter for one decision.

\subsection{Bad Experiences}\label{sec:bad}

A \emph{bad experience} is a logged update whose transition carries a safety or task event. Penalizing such events is a common way to encode safety in the reward~\cite{garcia2015safe}.
The driving events are listed in Table~\ref{tab:events}\rev{, whereas the} classic MDPs use the hazards \rev{specific to} each problem (Sec.~\ref{sec:classic_setup}).
\rev{Events alone determine the label,} regardless of the total reward or action \rev{source. For} example, entering a FrozenLake hole \rev{yields a reward of}~0.
We write $\mathcal{B}\subseteq\{1,\dots,T\}$ for the indices of \rev{a run's} bad experiences.

\begin{table}[t]
\centering
\caption{Driving events. A decision with several events takes the label listed first and, for minor hazards, incurs one penalty.}
\label{tab:events}
\footnotesize
\setlength{\tabcolsep}{3pt}
\begin{tabular}{@{}l p{3.8cm} c c@{}}
\toprule
Event & Definition & Penalty & Terminal \\
\midrule
Collision & Contact with a vehicle or pedestrian & 100 & yes \\
Road departure & Vehicle centre off the road & 100 & yes \\
Failure to yield & 1.5\,m/s or faster within 4\,m of a pedestrian on or entering the path & 1 & no \\
Offside excursion & A body corner in the oncoming lane & 1 & no \\
Lane-edge contact & A body corner off the road while the centre is on it & 1 & no \\
Insufficient clearance & An object closer than 2\,m ahead or 1\,m beside & 1 & no \\
\bottomrule
\end{tabular}
\end{table}

\subsection{Counterfactual Policies by Exact Replay}\label{sec:replay}

Given a group $\mathcal{G}\subseteq\mathcal{B}$, we replay the logged updates in their original order, \rev{starting} from the run's initial table $Q_0$ and \rev{using} the logged step sizes $\alpha_t$, \rev{and recompute} the temporal-difference target of \eqref{eq:qlearning} from the replayed table.
Removal omits the updates in $\mathcal{G}$.
Neutralization applies them with $r_t$ replaced by the reward without the event penalty, while retaining the transition and \rev{the} terminal flag.
Replaying all updates \rev{yields} \Full{}, whose Q-table \rev{is identical to} the trained one.
\Skip{} omits all bad experiences ($\mathcal{G}=\mathcal{B}$), whereas \Neut{} retains them without their event penalties.
For a terminal event, \Neut{} \rev{preserves the termination} and thus separates the penalty from \rev{termination. For} a minor hazard, it \rev{removes only} the penalty.
In the curriculum, $Q_0$ retains all earlier training, so interventions affect only the current stage.
Replay is exact conditional on this table, the logged transitions and the logged step \rev{sizes. However,} it does not model how retraining would change exploration and teacher intervention. Because a pair's first update has $\alpha_t=1$ and overwrites the initial value, skipping \rev{this update} leaves part of that initial value.

\subsection{Global Explanation by Counterfactual Policies}\label{sec:global}

The global explanation asks whether behavior changes when the bad experiences or their penalties are removed. Such a change \rev{indicates} that the bad experiences carry decision-relevant information, the prerequisite for the local explanation.
We compare \Full, \Skip{} and \Neut{} in closed-loop rollouts from the same start poses and report the goal rate.

\subsection{Local Explanation by Bad Experiences}\label{sec:local}

The local explanation answers ``why did the agent choose action A rather than action B?''~\cite{miller2019explanation}, where A is the trained agent's action (the fact) and B is the action of a failing counterfactual policy (the foil).
The foil is thus supplied by the failing policy rather than by a \rev{questioner, although} a foil chosen by a developer can be explained by the same search.
For each start \rev{from which} \Full{} reaches the goal and a policy C (\Skip{} or \Neut) fails, we \rev{select} the last of C's final ten decisions (2\,s) at which \Full{} would act \rev{differently, termed} the \emph{explained decision}.
We then search for the smallest group of bad experiences whose removal (if C is \Skip) or neutralization (if C is \Neut) makes A no longer greedy, and record whether \Full{} then chooses B.
The group is removed from the training log and the whole log is replayed; the test rollout only selects the state to explain. Unlike \Skip{} and \Neut, which change all bad experiences of a stage, the group changes only a few, so the table replayed without it is otherwise identical to \Full's.
\rev{The candidate} groups are the bad experiences at that state, grouped by action and event or by training episode, \rev{and} the same groups at the states reached under A or B, so the group is minimal only among these candidates.
\rev{This} smallest group \rev{constitutes} \emph{the explaining bad experiences}. The explanation reports $Q(\mathrm{A})$ and $Q(\mathrm{B})$ with and without them, together with the training episode, step and event of each \rev{experience}, for example ``episode 435, \act{straight + throttle}, road departure.''

\subsection{Tests of the Explanations}\label{sec:tests}

Following intervention-based evaluation of RL explanations~\cite{atrey2020exploratory}, we test three properties, each guarding against a different way an explanation can mislead.
\emph{Foil reproduction} tests whether \Full, without the explaining group, chooses the foil B rather than a third action, and how many bad experiences \rev{this change requires}.
\emph{Specificity} tests whether removing the same number of other bad experiences, chosen at random away from the explained decision, leaves the decision \rev{unchanged. As a control,} we also remove ordinary, unpenalized experiences of B at the same state.
\emph{Behavioral relevance} tests whether the changed decision \rev{alters driving. We} drive in closed loop\cite{irfan2026closedloop} and hand control from C to \Full{} at the explained decision.

\section{Experiments and Results}\label{sec:experiments}

The code, the training logs and the six classic MDP problems are available in an anonymous repository.\footnote{\url{https://anonymous.4open.science/r/DBEAST/}}

\subsection{Classic MDP Setup}\label{sec:classic_setup}

Six small, fully observable MDPs illustrate the \rev{mechanism, each} through one training run and one example \rev{case}.
The bad experiences are a hole in GridWorld ($6\times6$, nonterminal) and in FrozenLake (terminal, reward~0), an illegal pickup or drop-off in Taxi, a cliff fall in CliffWalking (reset to the start), entering an added penalty zone in MountainCar, and an off-target cell or an added penalty zone in GeoSteering ($\gamma=0.5$).
Taxi, FrozenLake and CliffWalking follow their Gymnasium definitions~\cite{towers2024gymnasium}.
FrozenLake \rev{serves as} the control case, \rev{because} neutralizing a zero reward \rev{has no effect}.
The agents use tabular Q-learning with $Q_0=0$ and 500 training episodes, and policies are evaluated greedily with the original rewards.

\subsection{Driving Setup in CARLA}\label{sec:carla_setup}

\begin{figure}[t]
\centering
\includegraphics[width=\columnwidth]{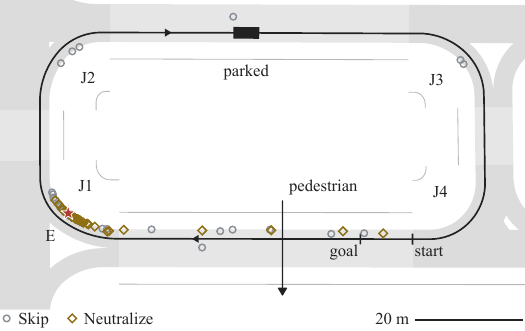}
\caption{Full-task route with the explained decisions before the failures of \Skip{} and \Neut. The star marks example E.}
\label{fig:map}
\end{figure}

\textit{Task and state representation.}
The agent drives along fixed routes in CARLA~0.9.15~\cite{dosovitskiy2017carla} (Town10HD). The full task is a 232\,m loop with four right-turn junctions, one parked vehicle and one pedestrian (Fig.~\ref{fig:map})\rev{, and the} pedestrian crosses in 80\% of the training episodes.
Every 0.2\,s\rev{,} the agent directly controls steering and pedals through nine joint actions, \rev{each} combining left, straight or right steering with throttle, coast or brake.
The policy uses seven discretized \rev{features, namely} speed, steering state, lateral offset, bearing to the route, object type, object gap, and whether the object is on the ego path.
It learns a tabular Q-function with \eqref{eq:qlearning} and $\gamma=0.98$.
The reward is \rev{the} progress in meters minus 0.1 per decision, with a cost of 0.4 for unnecessary standstill and a goal bonus of 50. The event penalties follow Table~\ref{tab:events}\rev{, and the} standstill cost is not an event penalty. Since a terminal event occurs at most once per episode, most bad experiences are minor hazards.

\textit{Curriculum.}
Training follows a curriculum~\cite{narvekar2020curriculum} through \rev{six stages, namely} straight road, turn, loop, parked vehicle, pedestrian, and the full task, which combines the loop with both objects. \rev{The} episode caps are 150, 250, 500, 200, 250 and 600 \rev{decisions, respectively}.
Each stage starts from the Q-table of the previous stage and passes its gate when the greedy policy reaches the goal from at least 16 of 20 perturbed start poses.
The curriculum \rev{combines} the learner's $\varepsilon$-greedy episodes with demonstrations from a rule-based teacher. In the pedestrian stage and \rev{the} full task, the teacher can also \rev{intervene} when a yield rule \rev{is triggered}, similar to human intervention during training~\cite{saunders2018trial}.

The full-task entry in Table~\ref{tab:carla} covers \rev{only the} final 1,200 training episodes \rev{of that stage}; earlier full-task episodes \rev{form} part of its initial table, \rev{as do all} earlier stages.

\textit{Test starts.}
The explanations are \rev{evaluated} from 40 start poses\rev{ not used} in training, with lateral and heading perturbations of $\pm0.5$\,m and $\pm0.1$\,rad\rev{, called} test starts.
Rollouts are greedy with the same rollout for all policies, and replay reproduces the trained Q-table and all recorded decisions exactly. Differences from \Full{} are \rev{assessed} with exact McNemar tests, Holm-corrected over the 12 comparisons.

% (results continue the Experiments and Results section)

\subsection{Classic MDP Results}\label{sec:classic_results}

The six cases, whose greedy rollouts and explained states are in the repository, illustrate how bad experiences affect decisions and routes.
First, removing or neutralizing bad experiences can change a greedy decision.
In CliffWalking, \Skip{} leaves $Q(\text{start},\text{Right})$ at its initial value (0.00, \rev{versus} $-109.17$ under \Full) and falls \rev{off} the cliff until timeout, whereas \Neut{} \rev{still chooses} Up. In GridWorld, \Neut{} reverses the start margin ($Q(\text{Up})/Q(\text{Down})$ of 7.77/8.83, \rev{versus} 4.44/$-1.17$ under \Full) and crosses one hole.
Second, a decision can change without \rev{altering} the behavior \rev{along} the route.
In Taxi, \Skip{} prefers an illegal pickup at an off-route state, but all three policies follow the same route. In the FrozenLake control, \Neut{} \rev{is identical to} \Full{} (holes carry \rev{a} reward \rev{of}~0), \rev{whereas} \Skip{} changes the greedy action in one state and the goal rate (7.5\% \rev{versus} 11.0\% of 200 evaluation rollouts).
In MountainCar, the explained decision at the start is the same under all three policies, but the rollouts diverge later, and \Skip{} and \Neut{} each enter the penalty zone twice before reaching the goal.
In GeoSteering, \Skip{} avoids the off-target cell but \rev{obtains} a lower discounted return ($1.2\times10^{-4}$ \rev{versus} $1.9\times10^{-4}$ for \Full)\rev{. In this case,} fewer penalties did not \rev{yield} a higher return.
% Taxi, FrozenLake and CliffWalking come from a one-seed demonstration; GeoSteering comes from a study stopped
% at 3,603 of 18,750 planned trainings; the MountainCar case was chosen among runs in which Full reaches the goal.

\subsection{Global Explanation}\label{sec:global_results}

\begin{table}[t]
\centering
\caption{Goals reached by stage. Ep.: episodes. Gate: goals of \Full{} from 20 gate starts. Bold means significant difference from \Full.}
\label{tab:carla}
\footnotesize
\setlength{\tabcolsep}{2pt}
\begin{tabular}{@{}l r r c c c c@{}}
\toprule
 & & & Gate & \multicolumn{3}{c}{Test starts (of 40)} \\
\cmidrule(lr){5-7}
Stage & Ep. & Bad / all updates & (of 20) & \Full & \Skip & \Neut \\
\midrule
Straight & 150 & 2,742 / 14,461 & 20 & 40 & 40 & 40 \\
Turn & 900 & 20,001 / 78,448 & 18 & 38 & 40 & 40 \\
Loop & 400 & 11,809 / 60,280 & 20 & 40 & \textbf{27} & \textbf{9} \\
Parked vehicle & 800 & 24,227 / 75,980 & 17 & 38 & 39 & 38 \\
Pedestrian & 600 & 7,346 / 53,667 & 19 & 34 & 32 & \textbf{10} \\
Full task$^\ast$ & 1,200 & 75,614 / 344,627 & 17 & 33 & \textbf{0} & \textbf{0} \\
\bottomrule
\multicolumn{7}{@{}p{0.97\columnwidth}@{}}{\scriptsize $^\ast$Final 1,200 episodes, whose bad experiences alone are skipped or neutralized.}
\end{tabular}
\end{table}

Removing the bad experiences of a stage lowered the goal rate in the loop and the full task, and removing only their penalties lowered it in the loop, pedestrian and full-task stages (Table~\ref{tab:carla}).
In the straight, turn and parked-vehicle stages, the three policies \rev{differed} by at most two of 40 starts.
Most failures were road departures, \rev{including} all 40 failures of \Neut{} in the full task.

\Neut{} reached the goal less often than \Skip{} in the loop (9 versus 27) and \rev{the} pedestrian stage (10 versus 32), and equally often in the full task (0).
\rev{Retaining} the experiences without their penalties thus lowered the goal rate at least as much as removing them, which suggests that the penalties \rev{deter} the agent from the penalized actions.
At the gate start in Fig.~\ref{fig:framework}(e), \Full{} stops for the pedestrian and reaches the goal, whereas \Neut{} collides with the pedestrian.

\subsection{Local Explanation of the Decision Before a Failure}\label{sec:local_results}

The local explanation answers the question posed in Sec.~\ref{sec:intro}.
In the loop, pedestrian and full-task stages, there are 143 cases in which \Full{} reaches the goal from a test start and \Skip{} or \Neut{} fails.
In 140 of \rev{these cases}, one of the \rev{failing policy's} last ten decisions is taken where \Full{} would act differently. Counting each combination of stage, encoded state and failing policy \rev{only} once \rev{yields} 47 explained decisions. Most lie inside a junction, and most are followed by a road departure or a lane-edge contact (Fig.~\ref{fig:map}). \Full{} itself passed the explained state in only 38 of the 140 cases.

\begin{figure}[t]
\centering
\includegraphics[width=\columnwidth]{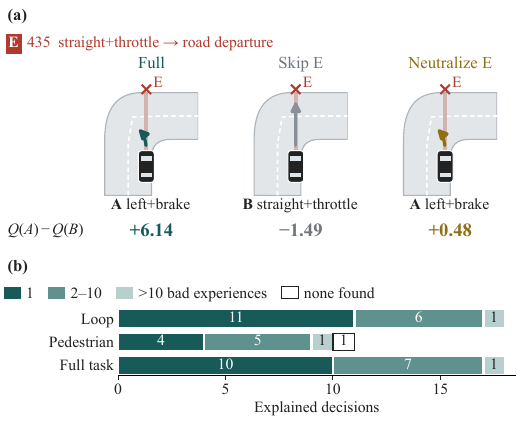}
\caption{(a)~Explanation of one decision, namely why the agent chose left+brake (A) rather than straight+throttle (B) before a road departure of \Skip. (b)~Size of the explaining group, the smallest group of bad experiences whose removal or neutralization makes \Full{} choose B, by stage.}
\label{fig:decisions}
\end{figure}

For all 47 explained decisions\rev{,} the search found a group that changed \Full's choice, and for 46 \rev{of them,} a group that made \Full{} choose B (Fig.~\ref{fig:decisions}(b)). For 25 \rev{decisions}, a single bad experience sufficed.
Removing the same number of other bad experiences at random never changed the choice, as expected for a tabular agent.
\rev{By contrast, removing} ordinary experiences of B at the same state changed \rev{the choice} for 6 of the 26 decisions that had such \rev{experiences, so bad experiences are} not always the only experiences \rev{underlying} these decisions.
In 29 of the 46 groups, the explaining bad experiences are earlier \rev{attempts} of B in the same state that ended in a lane-edge contact, an offside excursion or a road \rev{departure. In 15 groups,} they lie in the state \rev{to which} B \rev{leads}, and in 2, they \rev{comprise} all bad experiences at the explained state.

Example E (Fig.~\ref{fig:decisions}(a)) \rev{illustrates} the answer \rev{that} the method \rev{provides}.
Why did \Full{} choose \act{left + brake} rather than \act{straight + throttle}? \rev{The answer is that} in training episode 435, \act{straight + throttle} in this state led to a road departure, which lowered its value from 41.65 to 32.53.
With E, \Full{} prefers \act{left + brake} by 6.14. Skipping only E makes \act{straight + throttle} the greedy action ($-1.49$), whereas neutralizing only E's penalty \rev{retains} \act{left + brake} (0.48).
Relative to skipping E, \rev{retaining} E without its penalty raises the margin by 1.97 through its non-event reward and the \rev{termination} of the episode, and the penalty raises it by a further 5.66.

\textit{Closed-loop handover.}
\rev{Policy} C drove ten failing rollouts up to the explained decision, \rev{after which} \Full{} took control, either with its trained table or with the table replayed without the explaining bad experiences.
The ten cases were the first \rev{gate starts} with an explaining \rev{group, and these starts} are not held \rev{out. One} run deviated before the handover and was excluded. \Full{} reached the goal in 3 of the 9 valid cases and, without the explaining bad experiences, failed in all \rev{nine, as C did,} with the same event and episode length.
In the other six cases, the \rev{explained} decision \rev{occurred} one to four steps before the road departure, too late for \Full{} to recover.

\section{Discussion and Limitations}\label{sec:discussion}

\textit{What the explanations show.}
For this agent, decisions can be traced to its bad experiences, with two qualifications. \rev{First, the} global effect appears in three of the six stages rather than throughout the \rev{curriculum. Second,} some explained decisions also rest on ordinary experiences.
Hiraoka et al.~\cite{hiraoka2025influential} remove experiences whose influence on the agent is \rev{negative, and our results complement this approach}. A penalized experience can still support the learned policy, so the sign of an experience's reward does not indicate the sign of its influence.
Under \Neut{}, the penalized action is still updated with the non-event reward and may become preferred.
Under \Skip{}, it may retain its optimistic initial value, as in CliffWalking.

\textit{Scope of the local explanation.}
The explained state lies on the failing policy's path, which \Full{} itself rarely passed. The handover \rev{therefore} places \Full{} on that path, where it \rev{required} these experiences to recover.

\textit{Limitations.}
The counterfactuals show the effect of editing a fixed training log, not of retraining, which would also change exploration and teacher intervention.
Bad experiences \rev{constitute 14\%} to 32\% of the updates of a stage, and many full-task choices are near ties, with a median gap of 0.009 between the two highest action values on gate rollouts.
Without a size-matched removal of ordinary updates, part of the global effect may reflect the number of edited updates rather than their content, and the small explaining groups may partly reflect near ties.
Each stage was trained \rev{only} once, so the tests cover start poses for \rev{one} table rather than \rev{multiple} training runs, and the full-task comparison covers only \rev{the} final 1,200 episodes.
The handover test covers nine valid cases from one set of gate starts.
The seven driving features can map distinct physical situations to the same encoded state, and some explained decisions rest on \rev{limited data. For} 5 of the \rev{47,} the fact was never updated, and 4 are exact ties broken by action order.
Finally, the method \rev{remains tabular}. Replay is exact only for a Q-table over discrete encoded states and a finite action set, and function approximation would require approximate influence estimates with their own fidelity checks~\cite{koh2017influence,hiraoka2025influential}.

\section{Conclusion}\label{sec:conclusion}

We asked whether a driving agent's decisions can be explained by its bad experiences. Through exact replay of its fixed training log, we compared the agent trained with its bad \rev{experiences against replays} without them and with their penalties neutralized.
On the full driving task, \Full{} reached the goal from 33 of 40 test starts, whereas \Skip{} and \Neut{} reached it from none.
The local analysis linked 47 action choices before failures of \Skip{} and \Neut{} to small groups of bad experiences. When the trained agent took control at the explained decision, it reached the goal in 3 of 9 cases and, without those experiences, failed in all of them.
For this agent, decisions could thus be traced to recorded safety and task events, and the method \rev{quantifies} the contribution of their penalties within a fixed training log.
Future work will extend the method to explicitly partially observable settings (POMDPs) and to continuous state and action spaces, where the tabular results can serve as an exact reference for approximate attribution.

\bibliographystyle{IEEEtran}
\bibliography{references}

\end{document}